\documentclass[10pt,twocolumn,letterpaper]{article}

\usepackage{cvpr}              

\usepackage{graphicx}
\usepackage{amsmath}
\usepackage{amssymb}
\usepackage{booktabs}
\usepackage[numbers,sort,compress]{natbib}
\usepackage{bm}
\usepackage{pifont}
\usepackage{multirow}
\usepackage[pagebackref,breaklinks,colorlinks]{hyperref}

\usepackage[capitalize]{cleveref}
\crefname{section}{Sec.}{Secs.}
\Crefname{section}{Section}{Sections}
\Crefname{table}{Table}{Tables}
\crefname{table}{Tab.}{Tabs.}

\def\cvprPaperID{5675} 
\def\confName{CVPR}
\def\confYear{2023}

\begin{document}

\title{ACR: Attention Collaboration-based Regressor \\ for Arbitrary Two-Hand Reconstruction}
\author{Zhengdi Yu$^{1,2}$,  \hspace{1 mm} Shaoli Huang$^{1}$\thanks{Corresponding author.}, \hspace{1 mm} Chen Fang$^{1}$,\hspace{1 mm} Toby P. Breckon $^{2}$,\hspace{1 mm} Jue Wang\\
$^{1}$Tencent AI Lab \qquad $^{2}$Durham University\\
{\tt\small \{zhengdiyu,shaolihuang,fcfang\}@tencent.com} \qquad {\tt\small toby.breckon@durham.com} \qquad {\tt\small arphid@gmail.com}}


\maketitle

\begin{abstract}

Reconstructing two hands from monocular RGB images is challenging due to frequent occlusion and mutual confusion. 
Existing methods mainly learn an entangled representation to encode two interacting hands, which are incredibly fragile to impaired interaction, such as truncated hands, separate hands, or external occlusion. 
This paper presents ACR (Attention Collaboration-based Regressor), which makes the first attempt to reconstruct hands in arbitrary scenarios. 
To achieve this, ACR explicitly mitigates interdependencies between hands and between parts by leveraging center and part-based attention for feature extraction. 
However, reducing interdependence helps release the input constraint while weakening the mutual reasoning about reconstructing the interacting hands. 
Thus, based on center attention, ACR also learns cross-hand prior that handle the interacting hands better. 
We evaluate our method on various types of hand reconstruction datasets. 
Our method significantly outperforms the best interacting-hand approaches on the InterHand2.6M dataset while yielding comparable performance with the state-of-the-art single-hand methods on the FreiHand dataset. 
More qualitative results on in-the-wild and hand-object interaction datasets and web images/videos further demonstrate the effectiveness of our approach for arbitrary hand reconstruction. Our code is available at \href{https://github.com/ZhengdiYu/Arbitrary-Hands-3D-Reconstruction}{https://github.com/ZhengdiYu/Arbitrary-Hands-3D-Reconstruction}.

\end{abstract}

\vspace{-20pt}
\section{Introduction}
\label{sec:intro}
3D hand pose and shape reconstruction based on a single RGB camera plays an essential role in various emerging applications, such as augmented and virtual reality (AR/VR),  human-computer interaction, 3D character animation for movies and games, etc. However, this task is highly challenging due to limited labeled data, occlusion, depth ambiguity, etc. Earlier attempts~\cite{boukhayma20193d,zhang2019end,baek2019pushing,zhou2020monocular} level down the problem difficulty and focus on single-hand reconstruction. These methods started from exploring weakly-supervised learning paradigms~\cite{boukhayma20193d} to designing more advanced network models~\cite{tang2021towards}. Although single-hand approaches can be extended to reconstruct two hands, they generally ignore the inter-occlusion and confusion issues, thus failing to handle two interacting hands.
\begin{figure}
  \centering
  \includegraphics[width=1.0\linewidth]{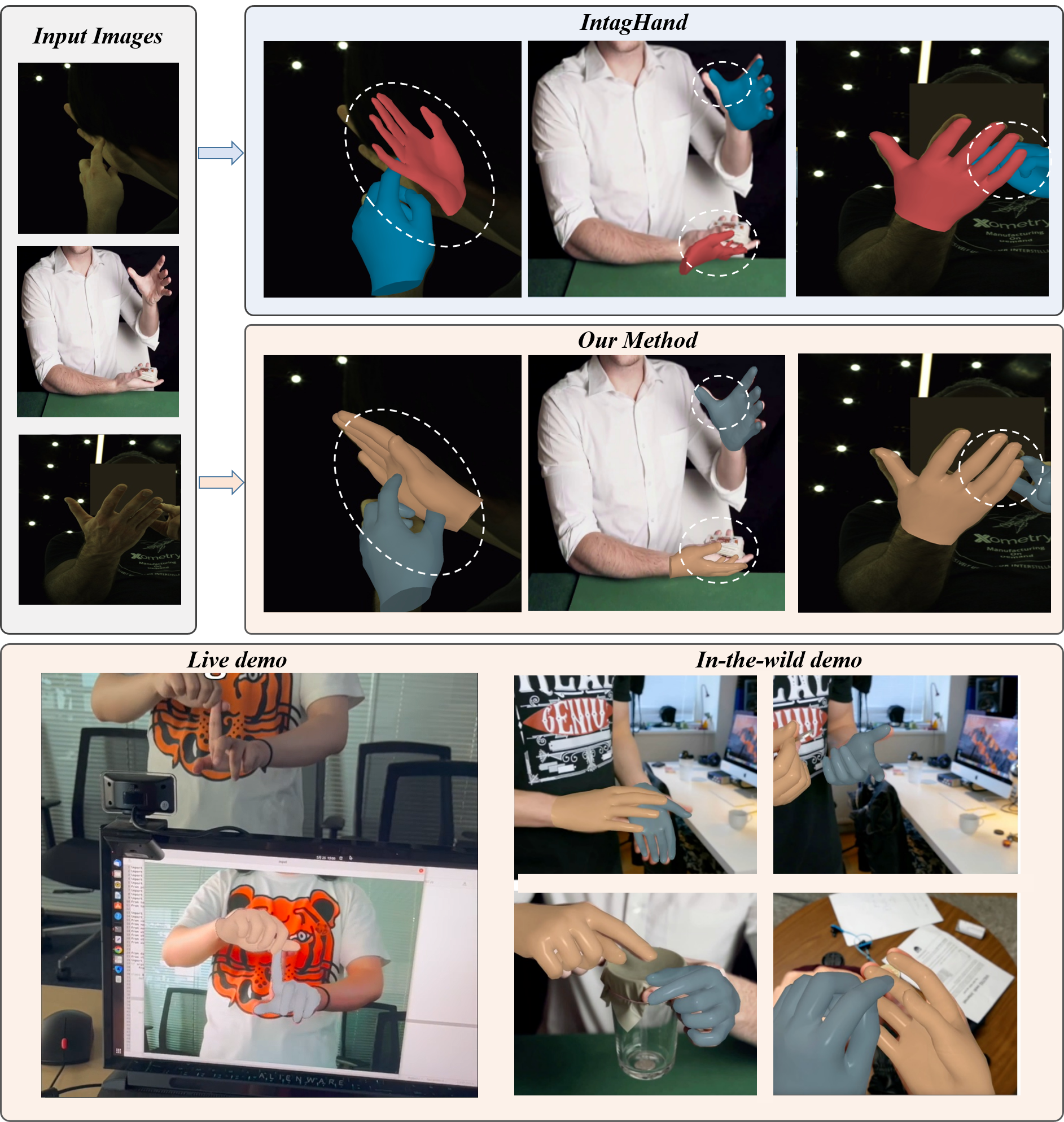}
  \caption{Given a monocular RGB image, our method makes the first attempt to reconstruct hands under arbitrary scenarios by representation disentanglement and interaction mutual reasoning while the previous state-of-the-art method IntagHand \cite{Li2022intaghand} failed.}
  \label{fig:open}
\end{figure}

To this end, recent research has shifted toward reconstructing two interacting hands.  \citet{wang2020rgb2hands} extract multi-source complementary information to reconstruct two interacting hands simultaneously. \citet{rong2021monocular} and \citet{zhang2021interacting}  first obtain initial prediction and stack intermediate results together to refine two-hand reconstruction. The latest work \cite{Li2022intaghand} gathers pyramid features and two-hand features as input for a GCN-based network that regresses two interacting hands unitedly. These methods share the same principle: treating two hands as an integral and learning a unified feature to ultimately refine or regress the interacting-hand model. The strategy delivers the advantage of explicitly capturing the hands' correlation but inevitably introduces the input constraint of two hands. This limitation also makes the methods particularly vulnerable and easily fail to handle inputs containing imperfect hand interactions, including truncation or external occlusions.

This paper takes the first step toward reconstructing two hands in arbitrary scenarios. Our first key insight is leveraging center and part attention to mitigate interdependencies between hands and between parts to release the input constraint and eliminate the prediction sensitivity to a small occluded or truncated part. To this end, we propose Attention Collaboration-based Regressor (ACR). Specifically, it comprises two essential ingredients: Attention Encoder (AE) and Attention Collaboration-based Feature Aggregator (ACFA). The former learns hand-center and per-part attention maps with a cross-hand prior map, allowing the network to know the visibility of both hands and each part before the hand regression. The latter exploits the hand-center and per-part attention to extract global and local features as a collaborative representation for regressing each hand independently and subsequently enhance the interaction modeling by cross-hand prior reasoning with an interaction field. In contrast to the existing method, our method provides more advantages, such as hand detector free. Furthermore, experiments show that ACR achieves lower error on the InterHand2.6M dataset than the state-of-the-art interacting-hand methods, demonstrating its effectiveness in handling interaction challenges. Finally, results on in-the-wild images or video demos indicate that our approach is promising for real-world application with the powerful aggregated representation for arbitrary hands reconstruction.

Our key contributions are summarized as: \textbf{\textit{(1)}} we take \textbf{the first step toward reconstructing two hands at arbitrary scenarios}. \textbf{\textit{(2)}} We propose to \textbf{leverage both center and part based representation to mitigate interdependencies between hands and between parts} and release the input constraint. \textbf{\textit{(3)}} In terms of modeling for interacting hands, we \textbf{propose a cross-hand prior reasoning module with an interaction field} to adjust the dependency strength. \textbf{\textit{(4)}} Our method \textbf{outperforms existing state-of-the-art approaches significantly on the InterHand2.6M benchmark}. Furthermore, ACR is the most practical method for various in-the-wild application scenes among all the prior arts of hand reconstruction.

\section{Related Work}
\noindent
\textbf{Single-Hand Reconstruction:} 
Hand pose and shape reconstruction from monocular images has rapidly progressed thanks to the development of the 3D hand parameterized model (e.g., MANO \cite{romero2017embodied} and DeepHandMesh \cite{moon2020deephandmesh}). However, hand-mesh annotations are expensive and difficult to acquire, which constitutes the main obstacle for this task. Existing works~\cite{boukhayma20193d,zhang2019end,baek2019pushing,zhou2020monocular} tackled the issue mainly by exploiting weakly-supervised learning paradigms or synthesizing pseudo data. For example, \citet{boukhayma20193d}  utilized 2D/3D keypoints as weak supervision to guide MANO parameter regression. \citet{zhang2019end} and \citet{baek2019pushing} introduced segmentation masks as extra weak labels in training by employing a neural mesh renderer \cite{kato2018neural}. Rather than using standard 2D labels,  \citet{zhou2020monocular} leveraged motion capture data for weak supervision and proposed an inverse kinematics network to recover hand mesh from 3D keypoints. Generating pseudo data is another effective way to mitigate mesh-label scarcity. \citet{kulon2020weakly} adopted a parametric model-fitting approach to generate pseudo mesh ground truth, enabling fully-supervised training for mesh reconstruction. \citet{ge20193d} created a synthetic dataset by blending a rendered hand with a background image and further trained a Graph CNN-based method with full supervision. Recently, with the emergence of new hand pose and shape datasets (e.g., FreiHAND~\cite{zimmermann2019freihand}),  the latest work focused on developing more advanced network models or learning strategies to improve reconstruction accuracy. For example, \citet{moon2020i2l} proposed an image-to-lixel network that considers prediction uncertainty and maintains the spatial relationship. In addition, \citet{tang2021towards} proposed decoupling the hand-mesh reconstruction task into multiple stages to ensure finer reconstruction.  Though these approaches have steadily improved hand reconstruction from monocular images, they are dedicated to the solo hand and usually fail to work well on two-hand cases. In contrast, our method explicitly addresses the challenge of inter-hand occlusion and confusion and, therefore, can deal with two interacting hands.

\textbf{Two-Hand Reconstruction:} A straightforward way to deal with two-hand reconstruction is to locate each hand separately and then transform the task into single-hand reconstruction. This strategy is commonly adopted in full-body reconstruction frameworks \cite{joo2018total,xiang2019monocular,choutas2020monocular,zhang2021lightweight,zhou2021monocular,feng2021collaborative}. However, independently reconstructing two hands remains a failure in addressing interacting cases, as the closer hands usually inter-occlude and easily confuse the model prediction. Earlier works successfully dealt with hand interaction mainly relied on model fitting and multi-view or depth camera setup. For instance, \citet{taylor2017articulated} introduced a two-view RGBD capture system and presented an implicit model of hand geometry to facilitate model optimization. \citet{mueller2019real} simplified the system by only using a single depth camera. They further proposed a regression network to predict segmentation masks and vertex-to-pixel correspondences for pose and shape fitting. \citet{smith2020constraining} adopted a multi-view RGB camera system to compute keypoints and 3D scans for mesh fitting. To handle self-interaction and occlusions, they introduced a physically-based deformable model that improved the robustness of vision-based tracking algorithms. 

Recent interest has shifted to two-hand reconstruction based on a single RGB camera. \citet{wang2020rgb2hands} proposed a multi-task CNN that predicts multi-source complementary information from RGB images to reconstruct two interacting hands. \citet{rong2021monocular} introduced a two-stage framework that first obtained initial prediction and then performed factorized refinement to prevent producing colliding hands. Similarly, \citet{zhang2021interacting} predicted the initial pose and shape from deeper features and gradually refined the regression with lower-layer features. The latest work \cite{Li2022intaghand} introduced a GCN-based mesh regression network that leveraged pyramid features and learned implicit attention to address occlusion and interaction issues. However, these methods primarily treat two hands as an integral and implicitly learn an entangled representation to encode two-hand interaction. In contrast, our approach learns independent features for each hand and exploits attention-conditioned cross-hand prior with local and global cues to address interacting challenges collaboratively.  

\begin{figure*}[t]
\centering
    \includegraphics[width=0.99\linewidth]{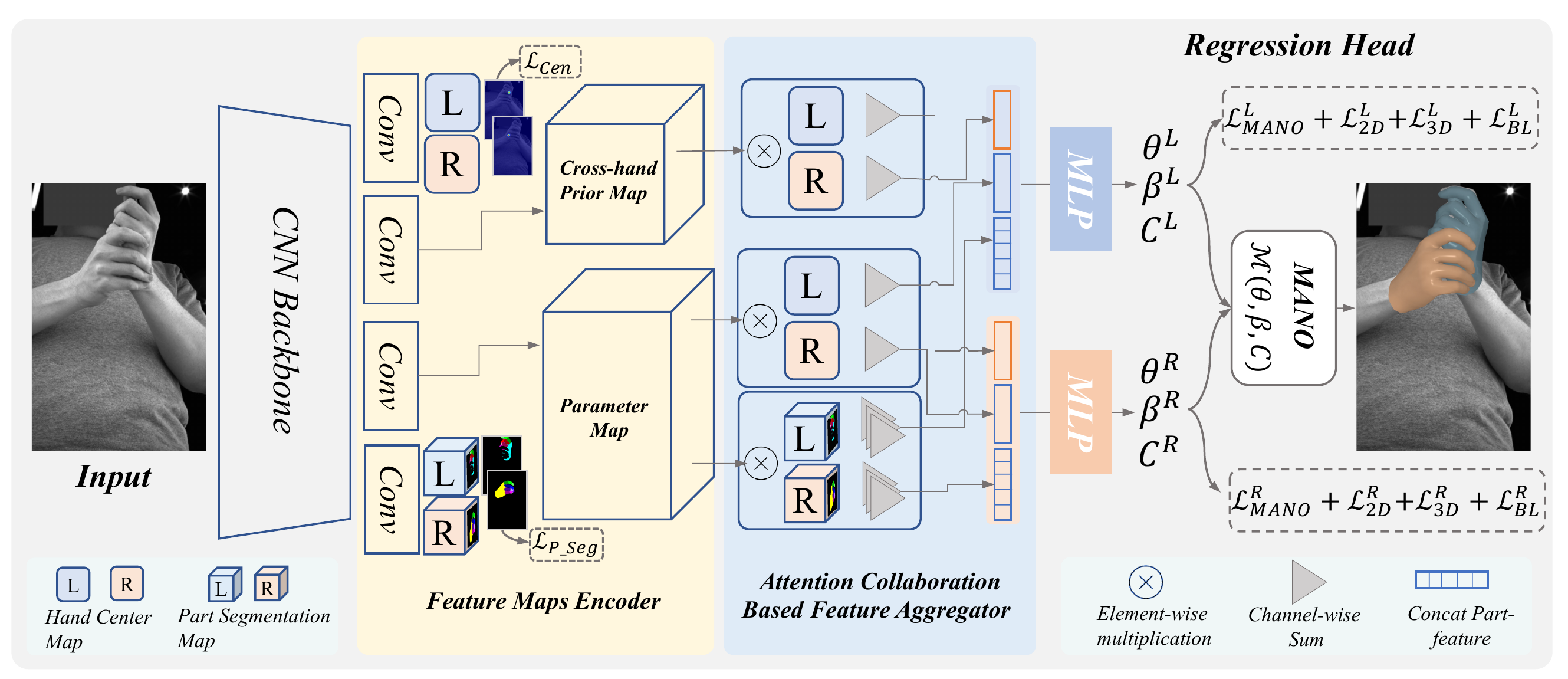}
    \caption{\textbf{ACR network architecture:} ACR takes a full-person image and uses a feature map encoder to extract hand-center maps, part-segmentation maps, cross-hand prior maps, and parameter maps. Subsequently, the feature aggregator generates the final feature for the hand model regression based on these feature maps. }
\label{fig:Overview}
\end{figure*}

\section{Methodology}\label{Methodology}


\noindent
Unlike existing works \cite{fan2021digit, boukhayma20193d, Li2022intaghand, lv2021handtailor, zb2017hand} that rely on an external detector to perform entangled bounding-box-level representation learning. Fig. \ref{fig:Overview} presents the overview of our method ACR. Given a single RGB image $\mathbf{I}$ as input, ACR outputs 4 maps, which are Cross-hand Prior map, Parameter map, Hand Center map, and Part Segmentation map. Based on Parameter map, which predicts weak-perspective camera parameters and MANO parameters for both left hand and right hand at each pixel, ACR then leverages three types of pixel-level representations for attention aggregation from Parameter map. First, ACR explicitly mitigates inter-dependencies between hands and between parts by leveraging center and part-based representation for feature extraction using part-based attention. Moreover, ACR also learns a cross-hand prior for handling the interacting hands better with our third Cross-hand Prior map. Finally, after aggregating the representations, we feed estimated parameters \bm{$F_{out}$} to MANO \cite{romero2017embodied} model to generate hand meshes.

\subsection{Preliminaries: Hand Mesh Representation}
\noindent
We use a parametric model MANO \cite{romero2017embodied} to represent hand, which contains a pose parameter $ \theta \in \mathbb{R}^{16 \times 3}$ and a shape parameter $ \beta \in \mathbb{R}^{10}$. We utilize 6D representations \cite{zhou2019continuity} to present our hand pose as $\theta \in \mathbb{R}^{16 \times 6}$. The final hand mesh $M$ could be reconstructed via a differentiable MANO model: $ M = W(\beta, \theta)$. Subsequently, 3D joints $J_{3D} \in \mathbb{R}^{21 \times 3}$ can be retrieved from the mesh: $\hat{J_{3D}} =  RM$, where R is a pre-trained linear regressor and $M \in \mathbb{R}^{778 \times 3}$.

\subsection{Representations of Attention Encoder}
\noindent
In this section, we will present the details of each output map or AE (Attention Encoder) module and their representations as shown in Fig. \ref{fig:Overview}. Given a monocular RGB image, we first extract a dense feature map $F \in \mathbb{R}^{C \times H \times W}$ through our CNN backbone. ACR then leverages three types of pixel-level representations for robust arbitrary hand representations disentanglement and mutual reasoning under complex interaction scenarios. For clarity, we denote the handedness by $\bm{h} \in \{L, R\}$.

\textbf{Parameter map:} $ M_{p} \in \mathbb{R}^{218 \times H \times W}$ can be divided into two maps for left hand and right hand separately, where the first 109 dimensions are used for left-hand feature aggregation and the rest for the right hand. For each of the map $ M_{p}^{h} \in \mathbb{R}^{109 \times H \times W}$. The 109 dimensions consist of two parts, MANO parameter $\theta \in \mathbb{R}^{16 \times 6}$, $ \beta \in \mathbb{R}^{10}$ and a set of weak-perspective camera parameters $(s, t_{x}, t_{y})$ that represents the scale and translation for the 2D projection of the individual hand on the image. This map serves as our \textbf{base} module for aggregated representation learning.

\textbf{Hand Center map:} $A_{c} \in \mathbb{R}^{2 \times H \times W}$ consists of two parts for left hand and right hand, which can be represented as $A_{c}^{h} \in \mathbb{R}^{1 \times H \times W}$. Each of the maps is rendered as a 2D Gaussian heatmap, where each pixel represents the probability of a hand center being located at this 2D position. The center is defined as the center of all the visible \textbf{MCP} joints, the joints that connect fingers with palm. For adaptive global representation learning, we generate heatmaps by adjusting the Gaussian kernel size K according to the bounding box size of the hand in data preparation for supervision (details in Supplementary Material). As the \textbf{first} representation of ACR, this map explicitly mitigates inter-dependencies between hands and serves as an attention mask for better global representation learning.

\textbf{Part Segmentation map:} $A_{p} \in \mathbb{R}^{33 \times H \times W}$ is learnt as a probabilistic segmentation volume. Each pixel on the volume is a channel of probability logits over 33 classes which consists of 1 background and 16 hand part classes for each hand corresponding to MANO model. Thus we have $A_{p}^{h} \in \mathbb{R}^{16 \times H \times W}$. We obtain the part segmentation mask obtained by rendering the ground truth MANO hand mesh using a differentiable neural renderer \cite{kato2018neural}. As the \textbf{second} representation of ACR, this map serves as an attention mask for part representation learning.

\textbf{Cross-hand Prior map:} $M_{c} \in \mathbb{R}^{218 \times H \times W}$ contains two maps, $M_{c}^{h} \in \mathbb{R}^{109 \times H \times W}$. It is split into two sets of parameters which are MANO parameter $\theta \in \mathbb{R}^{16 \times 6}$, $\beta \in \mathbb{R}^{10}$ and 3 camera parameters for cross hand \textbf{inverse} feature query. Empirically, the two hands' pose will be highly correlated when they are closely interacting within interaction field (\textbf{IF}), which is introduced in \ref{mutual_reasoning}. As our \textbf{third} representation, aggregating this module into our robustly disentangled representations is providing us the powerful mutual reasoning ability under severe interaction scenarios. 

\subsection{Robust Representation Disentanglement}\label{disentangle}
\noindent
Unlike all the existing approaches for interacting hands reconstruction \cite{Li2022intaghand, zhang2021interacting, Moon_2020_ECCV_InterHand2.6M}, which require that the input image must be fixed to \textbf{two closely} interacting hands and occupy the most region of the image, thus causing ambiguity and unnecessary input constraints as shown in Fig. \ref{fig:open}, our first step towards building \textbf{arbitrary} hands representation is - \textbf{disentanglement} by decomposing the ambiguous hand representations. Thanks to the powerful pixel-wise representation of Hand Center map, we are able to disentangle \textbf{inter}-hand dependency and build an explicitly separate feature representation for the two hands. However, when the two centers are getting too close, these feature representations could also be highly ambiguous. Subsequently, for better disentangled feature representation learning, inspired by \cite{ROMP}, we adopt a collision-aware center-based representation to further split the features of two hands by applying Eq. \ref{car}. When the two hands are too close to each other with a Euclidean distance $d$ smaller than $k_{L} + k_{R} + 1$. The new centers will be generated as:

\begin{equation}\label{car}
\begin{split}
\bm{\hat{C_{L}}}& = C_{L} + \alpha R, \quad \bm{\hat{C_{R}}} = C_{R} - \alpha R,\\
\quad \bm{R} &= \frac{k_{L} + k_{R} + 1 - d}{d}(C_{L}-C_{R})
\end{split}
\end{equation}
where $C_{L}, k_{L}$ and $C_{R}, k_{R}$ stand for two hand centers and their kernel size. $R$ means the repulsion vector from $C_{L}$ to $C_{R}$. In addition, $\alpha$ refers to an intensity coefficient to adjust the strength. Finally, the global representation $\bm{F_{g}^{h}} \in \mathbb{R}^{J * 6 + (10 + 3)}$, is extracted by combing Hand Center map $A_{c}$ with parameter map $M_{p}$ as:

\begin{equation}\label{global_rep}
F_{g}^{h} = f_{g}(\sigma(A_{c}^{h})\otimes M_{p}^{h})
\end{equation}
where $\sigma, \odot$ and $f_{g}$ are spatial softmax, pixel-wise multiply and a point-wise Multi-Layer Perceptron (MLP) layer separately, and $\bm{h} \in \{L, R\}$.

With such global feature representation $F_{g}$, we have successfully disentangled inter-dependency. However, having only such global representation will lead to instability under occlusion and losing the ability to recover details, due to the unnecessary \textbf{inner} dependency of each hand part. Subsequently, we need to further disentangle our representation utilizing our Part Segmentation map $A_{p}$ following \cite{kocabas2021pare}. For simplicity, we ignore the $\bm{h} \in \{L, R\}$ here, the two hands follow the same formulation as:
\begin{equation}\label{part_rep}
F_{p}^{(j,c)} = \sum_{h, w} \sigma(A^{j}_{p}) \odot M_{p}^{c},
\end{equation}
where $F_{p} \in \mathbb{R}^{J \times C}$ is final part representation and $F_{p}^{(j,c)}$ is its pixel at (j, c). $\odot$ is the Hadamard product. Thus, the part segmentation maps after spatial softmax normalization $\sigma$ are used as soft attention masks to aggregate features in $M_{p}^{c}$. We follow prior arts to implement a dot product based method by reshaping the tensor at first: $F_{p} = \sigma(A_{p}^{*})^{T}M_{p}^{*} $, where $M_{p}^{*} \in R^{HW \times C}$ and $A_{p}^{*} \in R^{HW \times J}$ are the parameter map $M_{p}$ and reshaped part segmentation $A_{p}$ without background mask. 
Finally, the global feature representation $F_{g}$ and part representation and $F_{p}$ are aggregated into our Robust Inter and Inner Disentangled Representation.

\begin{figure}
  \centering
  \includegraphics[width=1\linewidth]{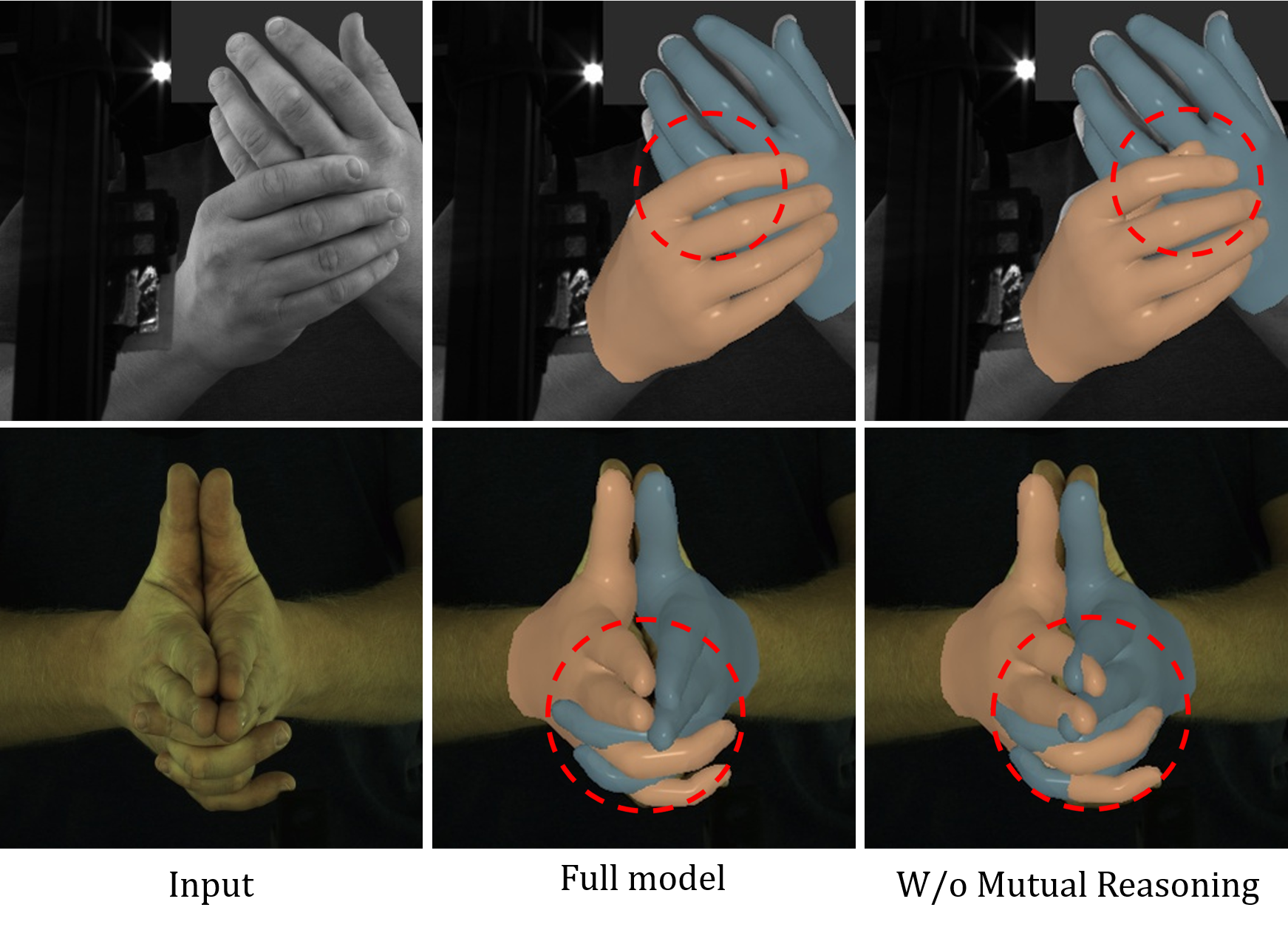}
  \caption{This compares our full model and a model without mutual reasoning module. It is shown that our mutual reasoning module explicitly helps to deduce and recover the correlation between closely interacting hands with less mutual occlusion.}
  \label{fig:qua}
\end{figure}
\subsection{Mutual Reasoning of Interaction}\label{mutual_reasoning}
\noindent
Despite the powerful disentangled representations, it has been explored that the states of two interacting hands are highly correlated \cite{zhang2021interacting, Li2022intaghand} when they are interacting closely. Simply disentangling inter and inner dependencies as the final representation will weaken the mutual reasoning about reconstructing the interacting hands.
Subsequently, we design a novel mutual reasoning strategy by reusing the center-based attention via a \textbf{inverse query}:

\begin{equation}\label{cross}
\centering
\begin{split}
F_{c}^{R\xrightarrow{}L} = f_{c}(\sigma(A_{c}^{R})\otimes M_{c}^{L}), \\
F_{c}^{L\xrightarrow{}R} = f_{c}(\sigma(A_{c}^{L})\otimes M_{c}^{R}),
\end{split}
\end{equation}
where $F_{c}^{R\xrightarrow{}L}$ is the left-hand prior representation that is deduced from right-hand attention and vice versa. $M_{c}$ is the output dense feature map from cross-hand-prior attention blocks, $A_{c}$ is our center based attention map, and L, R stand for left hand and right hand. $\sigma, \otimes$ and $f_{c}$ are spatial softmax, pixel-wise multiply and a point-wise MLP layer.

However, for two more distant hands or a single hand, the correlation between them should be mitigated or eliminated. Subsequently, we also propose a new mechanism, interaction field (\textbf{IF}) to adjust the dependency strength. Specifically, by first computing the Euclidean distance \bm{$d$} between the hands, when the two hands are too close to each other and entering the field of \textbf{IF}$ = \gamma(k_{L} + k_{R} + 1)$, where $\gamma$ is a field sensitivity scale, and the interaction intensity coefficient \bm{$\mathcal{\lambda}$} will be computed as:

\begin{align*}
\begin{split}
\bm{\lambda_(C_{L}, C_{R})}= \left \{
\begin{array}{ll}
                                0,                                 & d > IF\\
    \frac{\ IF - d}{d}||C_{L}-C_{R}||_{1},                    & d <= IF\\
\end{array}
\right.
\end{split}
\end{align*}
The interaction intensity coefficient \bm{$\mathcal{\lambda}$} helps our cross-hand prior representation to formulate an adaptive interaction field that can better model the correlations of two hands while keeping sensitive to close interaction and separation to avoid unnecessary feature entanglement. Finally, our final output self-adaptive robust representation could be represented as:
\begin{equation}\label{fianl_representation}
F_{out}^{h} = f_{out}(concat(F_{g}^{h}, F_{p}^{h*}, \lambda F_{c}^{\hat{h}\xrightarrow{}h}))
\end{equation}
where $f_{out}$ is point-wise MLP layers for regressing the final representation $F_{out}^{h} \in \mathbb{R}^{109}$, and $F_{c}^{h*} \in \mathbb{R}^{J*C}$ is reshaped part disentangled representation. Finally, the results are fed into MANO model to regress the final hand mesh. For simplicity, we represent the opposite hand by $\hat{h}$ in Eq. \ref{fianl_representation}.

\subsection{Loss Functions} 
\noindent
For training ACR with three types of powerful representation, our loss functions are divided into three groups, as demonstrated in Fig \ref{fig:Overview}. Specifically, ACR is supervised by the weighted sum of all loss items for both left hand and right hand: mesh recovery loss, center-based attention loss, and part-based attention loss.

\textbf{Center Attention Loss} can be treated as a segmentation problem, however, the Gaussian distribution on the image is a relatively small area and there is an imbalance between the positive and negative samples. Subsequently, we utilize focal loss \cite{lin2017focal} to supervise our center map regressor as:

\begin{equation}\label{center_map_loss}
\bm{\mathcal{L}_{c}} = \sum_{h \in \{L, R\}}f(A_{c}^{h}, \hat{A_{c}^{h}}),
\end{equation}
where $f$ is focal loss \cite{lin2017focal}, $h \in \{L, R\}$ means left hand and right hand, and $\hat{A_{c}^{h}}$ is the ground truth hand center map for hand type $h$. For simplicity, here we abbreviate the formulation of focal loss, which can be found in detail within the Supplementary Material.

\textbf{Part Attention Loss} is used to supervise our Part-based Representation learning. We only supervise this loss with CrossEntropy loss in the first 2 epochs and continue to train with other losses until it converges.
\begin{equation}\label{part_seg_loss}
\begin{split}
\bm{\mathcal{L}_{seg}} = \frac{1}{HW}\sum_{h, w}CrossEntropy(\sigma(A_{p}^{hw}), \hat{A_{p}^{hw}}),
\end{split}
\end{equation}
where $\hat{A_{p}}$ means GT part segmentation maps and $\hat{A_{p}^{hw}}$ is the ground truth class label at the location of (h,w). Different from our part soft attention mask, $A_{p}^{hw} \in \mathbb{R}^{33 \times 1 \times 1} $ here means the probabilistic segmentation volume at the pixel position of $(h, w)$ and $\sigma$ means softmax along channel dimension. We do not need to omit the background class here.

\textbf{Mesh Recovery Loss} is applied for each hand, thus we ignore the handedness $\bm{h} \in \{L, R\}$ here for simplicity. Finally, the loss for left hand and right hand will be summed into the total loss. Instead of relying on the ground truth vertex positions, which could cause degeneration in generalization ability, we decouple our mesh loss into 3 parts:
\vspace{-5pt}
\begin{equation}\label{mesh_recovery_loss}
\begin{split}
\bm{\mathcal{L}_{mesh}} = \bm{\mathcal{L}_{mano}} + \bm{\mathcal{L}_{joint}},
\end{split}
\end{equation}
where $\mathcal{L}_{mano}$ is the weighted sum of $L2$ loss of the MANO parameters $\theta$ and $\beta$, namely $w_{\theta}\mathcal{L}_{\theta} + w_{\beta}\mathcal{L}_{\beta}$:

\begin{equation}\label{mano_loss}
\begin{split}
\bm{\mathcal{L}_{\theta}} = w_{\theta}|| \theta - \hat{\theta}||^{2}_{2},  \quad \bm{\mathcal{L}_{\beta}} = w_{\beta} || \beta - \hat{\beta}||^{2}_{2},\\
\end{split}
\end{equation}
where$\mathcal{L}_{joint}$ is the weighted sum of $\mathcal{L}_{3D}$, $\mathcal{L}_{2D}$ and a bone length loss $\mathcal{L}_{bone}$ to present better geometric constraint to the reconstructed mesh, which is computed by $L2$ distance between $i^{th}$ ground truth bone length $\hat{b_{i}}$ and predicted length $b_{i}$:
\begin{equation}\label{joint_loss}
\begin{split}
\bm{\mathcal{{L}}}&_{\bm{3D}} = w_{j3d}\mathcal{L}_{MPJPE} + w_{paj3d}\mathcal{L}_{PA-MPJPE},\\
\bm{\mathcal{L}}&_{\bm{PJ2D}} = w_{pj2d} || PJ_{2D} - \hat{J_{2D}}||^{2}_{2},\\
\bm{\mathcal{L}}&_{\bm{bone}} =  \sum_{i} || b_{i} - \hat{b_{i}}||^{2}_{2},
\end{split}
\end{equation}
where $\mathcal{L}_{MPJPE}$ is the $L2$ loss between ground-truth 3D joints $\hat{J_{3D}}$ and predicted ones $J_{3D}$ retrieved from predicted mesh. $\mathcal{L}_{PA-MPJPE}$ is computed as the Procrustes-aligned mean per joint position error (PA-MPJPE). We do not supervise camera parameters directly, instead, the network adjusts the camera parameters by computing the $L2$ loss between ground truth $\hat{J_{2D}}$ and the projected 2d joints $PJ_{2D}$ retrieved by a weak-perspective camera: $PJ_{2D}$ as $x_{pj2d} = sx_{3D} + t_{x}, y_{pj2d} =  sy_{3d} + t_{y}$. Finally, to compute $\mathcal{L}_{mesh}$ as a weighted sum, we apply $w_{j3d}=200$, $w_{paj3d}=360$, $w_{pj2d}=400$, $w_{bl}=200$. For $\mathcal{L}_{mano}$, we use $w_{pose}=80$, $w_{shape}=10$ in our experiments.

\textbf{Total Loss} is the weighted sum of the described loss above and can be represented as:
\begin{equation}\label{total_loss}
\begin{split}
\bm{\mathcal{L}_{total}} = \bm{\mathcal{L}_{mesh}} + w_{c}\bm{\mathcal{L}_{c}} + w_{p}\bm{\mathcal{L}_{seg}},
\end{split}
\end{equation}
where $w_{c}=80$, $w_{p}=160$ and $\mathcal{L}_{mesh}$ is already a weighted sum. Each part is activated only when the corresponding ground truth is available. Finally, all of these losses are trained simultaneously in an end-to-end manner.
\begin{figure*}
    \includegraphics[width=0.99\linewidth]{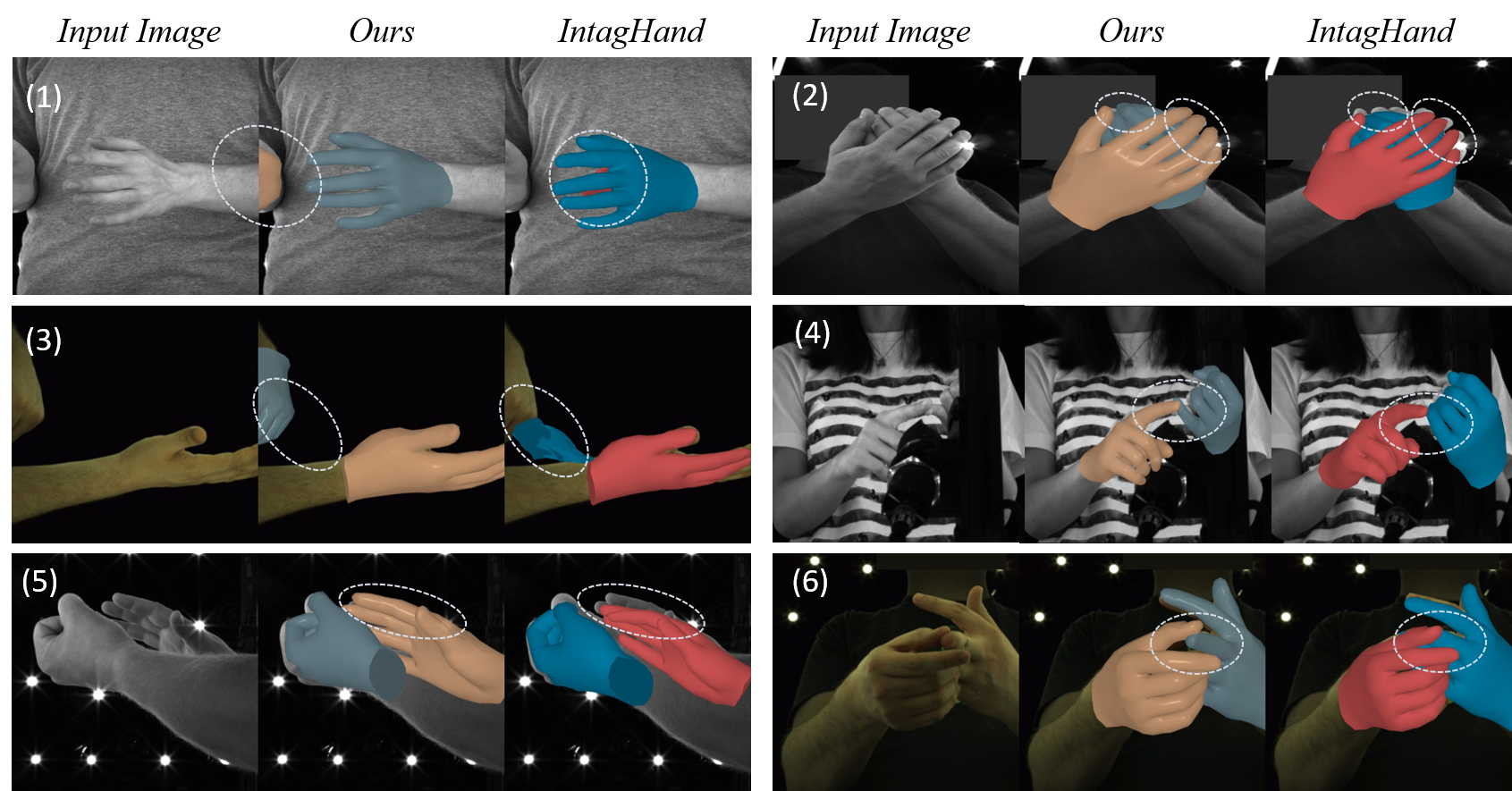}
    \caption{Qualitative comparison with on InterHand 2.6M test dataset. Our approach generates better results in two-hand reconstruction, particularly in challenging cases such as external occlusion (1), truncation (3-4), or bending one finger with another hand (6). More results can be found in the Supplementary Material.}
    \label{fig:interhand}
\end{figure*}

\section{Experiments}\label{Experiments}
\noindent
\textbf{Implementation details:} We implement our network based on PyTorch \cite{paszke2019pytorch}. For the backbone network, we have trained with both ResNet-50 \cite{he2016deep} and HRNet-W32 \cite{cheng2020higherhrnet}, for faster inference speed or better reconstruction results respectively. Unlike existing approaches that require a hand detector, our method can reconstruct arbitrary hands in an end-to-end manner. Furthermore, our method does not limit its input to two-hand. Given a monocular raw RGB image without cropping or detection, all the input raw images and segmentation maps are resized to $512\times512$ while keeping the same aspect ratio with zero padding, then we extract the feature maps $f \in R^{(C+2)\times H \times W}$ from the backbone network with CoordConv \cite{liu2018intriguing}. The feature maps are fed finally to four Conv blocks to produce the four maps for representation aggregation.

\textbf{Training:} For comparison on InterHand2.6M dataset, we train our model using Adam optimizer with a learning rate 5e-5 for eight epochs. We do not supervise $L_{seg}$ and $L_{MANO}$ when there is no MANO label valid because our ground truth segmentation is obtained from rendering ground truth MANO hand mesh using a neural renderer \cite{kato2018neural}. For all of our experiments, we initialized our network using the pre-trained backbone of HRNet-32W from \cite{fan2021digit} to speed up the training process. We train our network using 2 V100 GPUs with $batchsize$ of 64. The size of our backbone feature is $128\times128$ and the size of our 4 pixel-aligned output maps is $64 \times 64$. We applied random scale, rotation, flip, and colour jitter augmentation during training.

\textbf{Testing:} For all the experiments, if not specified, the backbone is HRNet-32W. For comparison with state-of-the-art, we use the full official test set for evaluation. The confidence threshold is set to 0.25 with a max detection number of one left hand and one right hand, as we only have one left hand and one right hand in all the training and testing sets.

\textbf{Evaluation Metrics:} To evaluate the accuracy of the two-hand reconstruction, we first report the mean per joint position error (MPJPE) and the Procrustes-aligned mean per joint position error (PA-MPJPE) in millimetres. Both errors are computed after joint root alignment following prior arts. We also studied the reconstruction accuracy of handshape by mean per-vertex position error (MPVPE) and the Procrustes-aligned mean per-vertex position error (PA-MPVPE) on the FreiHand dataset. Please see details of the metrics in supplementary materials.
\subsection{Datasets}
\noindent
\textbf{InterHand2.6M} \cite{Moon_2020_ECCV_InterHand2.6M} is the first one and the only publicly available dataset for two-hand interaction with accurate two-hand mesh annotations. This large-scale real-captured dataset, with both accurate human (H) and machine(M) 3D pose and mesh annotation, contains 1,361,062 frames for training and 849,160 frames for testing, and 380,125 for validation in total. These subsets are split into two parts: interacting hands (IH) and single hand (SH). We use the 5 FPS IH subset with H+M annotations for our experiments.

\noindent
\textbf{FreiHand} \cite{zimmermann2019freihand} is a single hand 3D pose estimation dataset. For each frame, it has MANO annotation and 3D keypoints annotation. There are 4$\times$32,560 frames for training and 3960 frames for evaluation and testing. The initial sequence with 32560 frames is captured with a green screen background, allowing background removal.

\begin{table*}
\centering

\resizebox{0.9\linewidth}{!}{
\begin{tabular}{c|c|c|c|c|c|c|c}
\toprule
\multicolumn{1}{l|}{} & \multicolumn{1}{l|}{extra info.} & \multicolumn{1}{l|}{MPJPE}& \multicolumn{1}{l|}{MPVPE} & \multicolumn{1}{l|}{IH MPJPE}  & \multicolumn{1}{l|}{IH MPVPE} & \multicolumn{1}{l|}{SH MPJPE}    & \multicolumn{1}{l}{SH MPVPE}  \\ 
\midrule\midrule
(-) Zimmermann et al.\cite{zb2017hand}   & Box & -  & -  & 36.36 & -  & -   & - \\
(-) Zhou et al.\cite{zhou2020monocular}  & Box   & -  &  -   & 23.48  & 23.89 & -   & -  \\
(-) Boukhayma et al.\cite{boukhayma20193d} & Box& - & - & 16.93 & 17.96 & - & - \\
(-) Spurr et al. \cite{spurr2018cross} & Box & - & -  & 15.40 &  - & -& -\\ 
\midrule
Moon et al. \cite{Moon_2020_ECCV_InterHand2.6M} & Box& 13.98 & -& 16.02 & -& 12.16  & - \\
Fan et al.  \cite{fan2021digit}  & Box& - & - & 14.27& -  & 11.32  & -\\
Zhang et al.  \cite{zhang2021interacting}   & Box & -  & -   & 13.48 & 13.95 & - & - \\ 
IntagHand \cite{Li2022intaghand} & Box & 9.95   & 10.29   & 10.27  & 10.53  & 9.67  & 9.91 \\
\textbf{Ours}    & \textbf{-}   & \begin{tabular}[c]{@{}c@{}}\textbf{8.09}\\\end{tabular} &\textbf{8.29}  & \begin{tabular}[c]{@{}c@{}}\textbf{9.08}\\\end{tabular}    & \textbf{9.31} & \begin{tabular}[c]{@{}c@{}}\textbf{6.85}\\\end{tabular} & \textbf{7.01}  \\
\midrule
IntagHand \cite{Li2022intaghand}& Box+scale & 9.18  & 9.42 & 9.40 & 9.68 & 9.0  & 9.18 \\
\textbf{Ours} & scale  & \textbf{7.41}  & \textbf{7.63}  & \begin{tabular}[c]{@{}c@{}}\textbf{8.41}\\\end{tabular} & \textbf{8.53} & \textbf{6.09}& \textbf{6.21} \\
\hline
\end{tabular}
}
\caption{Comparison with state-of-the-art on InterHand2.6M\cite{Moon_2020_ECCV_InterHand2.6M}. (-) means single hand reconstruction method. Except for our approach, all the others use ground-truth bounding boxes from the dataset. The single-hand results are taken from \cite{zhang2021interacting}. We report results on the official test split of the InterHand2.6M dataset for fair comparison. We noted that the reported result of IntagHand is obtained from a filtered test set. We, therefore, get the result on the standard test set by running its released code \cite{zhang2021interacting}.} 

\label{table:1}
\end{table*}
\subsection{Comparison to State-of-the-art Methods}
\noindent
\textbf{Results on InterHand2.6M and FeiHand datasets:} 
We first compare our method with single-hand and interacting-hand approaches on InterHand2.6M. We follow the official split to train our model, and we report results on the official test split of the InterHand2.6M dataset for a fair comparison. As the reported result of IntagHand is obtained from a filtered test set, we get the result on the standard test set by running its released code. Tab. \ref{table:1} presents comparison results on the \textbf{I}nteracting hands (IH MPJPE),  and \textbf{S}ingle hand (SH MPJPE) subset, and the full-set (MPJPE).  Not surprisingly, we can observe that single-hand methods generally perform poorly on the IH subset, as their method designs dedicate to single-hand input. Next, we perform a comparison with two state-of-the-art interacting-hand approaches  \cite{zhang2021interacting} and \cite{Li2022intaghand}. The first one adopted a refinement strategy that predicted the initial pose and shape from deeper features and gradually refined the regression with lower-layer features. The latter IntagHand incorporates pyramid features with GCN-based to learn implicit attention to address occlusion and interaction issues, while IntagHand is our concurrent work and outperforms  \cite{zhang2021interacting}. However, our proposed method constantly surpasses IntagHand without extra information needed. Specifically, our method obtained the lowest MPJPE of 8.49 on the IH subset, demonstrating its effectiveness in handling interacting hands. It also achieves a 6.91 MPJPE on the SH dataset that outperforms IntagHand by a large margin, showing our method remains superior on single-hand reconstruction.

\begin{figure}
  \centering
  \includegraphics[width=1\linewidth]{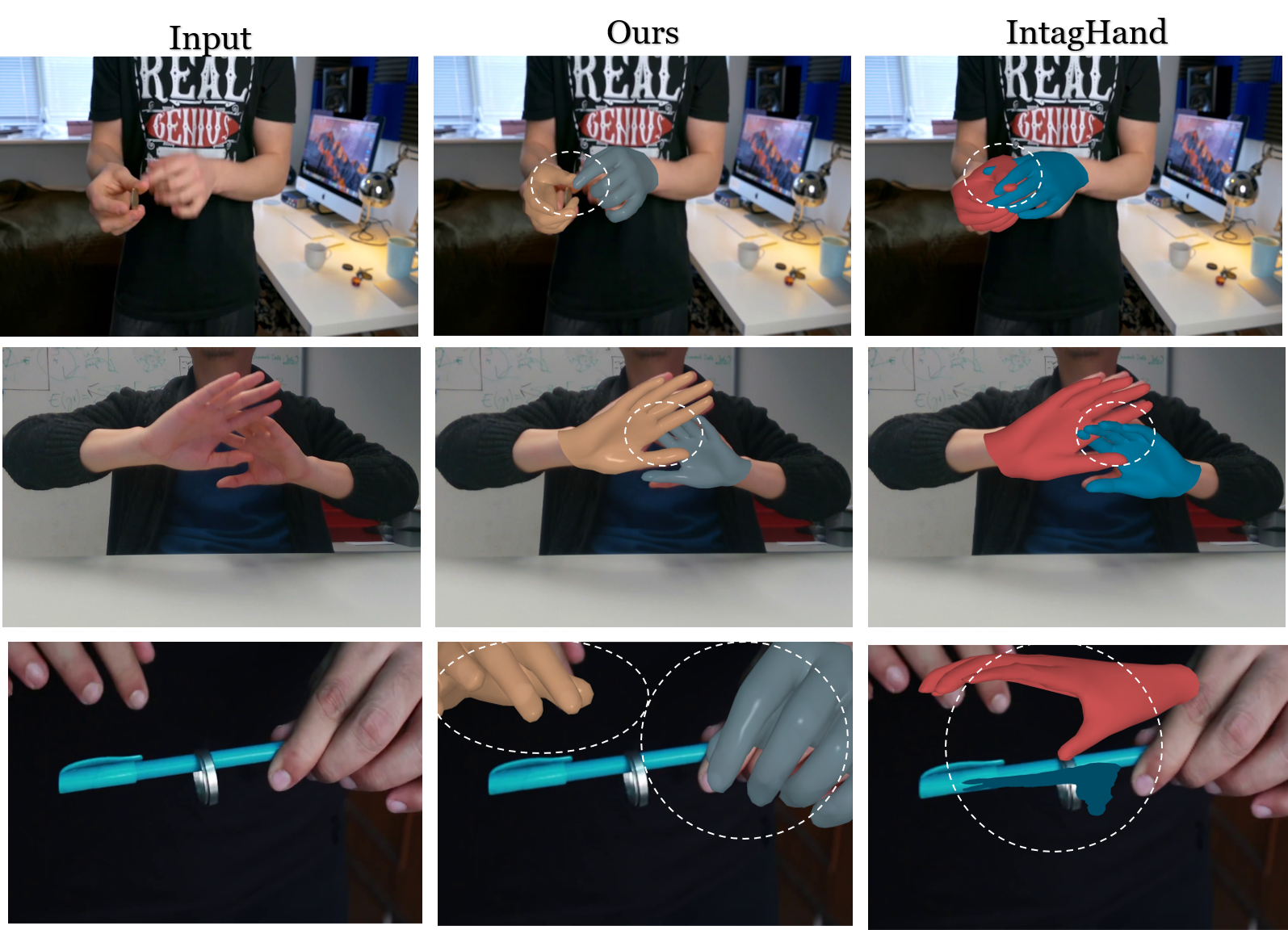}
  \caption{Qualitative comparison with IntagHand \cite{Li2022intaghand} on in-the-wild images.}
  \label{fig:qua2}
\end{figure}

\begin{table}
\centering
\scalebox{0.9}{
\begin{tabular}{ccc}
    \toprule
    
     Method &  PA-MPJPE &  PA-MPVPE \\
     \midrule  \midrule 
    Mesh Graphormer\cite{lin2021mesh} & 6 & 5.9   \\
    METRO\cite{lin2021end} & 6.8   & 6.7  \\
    I2L-MeshNet\cite{moon2020i2l}     & 7.4  & 7.6       \\
    HandTailor\cite{lv2021handtailor}   & 8.2   & 8.7       \\ 
    \hline
     ours   & 6.9       & 7.0       \\
    \hline
    \bottomrule  
\end{tabular}}
\caption{Comparison with state-of-the-art on FreiHand \cite{zimmermann2019freihand} Benchmark.} 
\label{table:fig5}
\end{table}


We also compare our method with single-hand methods on the single-hand dataset FreiHand \cite{zimmermann2019freihand}. We follow the official split to train and test our model on this dataset separately. As shown in Tab. \ref{table:fig5}, the transformer-based method achieves the best result. Nevertheless, our method obtains comparable performance to this state-of-the-art single-hand approach, revealing its potential to improve single-hand reconstruction.

\textbf{Qualitative Evaluation:} We previously demonstrated our method significantly outperforms IntagHand in quantitative experiments. To gain insight into this result, we conduct a qualitative comparison between these two methods. Interestingly, our approach generally produces better reconstruction results over IntagHand in challenging cases such as external occlusion and truncated hands. Fig. \ref{fig:interhand} shows some examples of these cases. This result indicates that our method for two-hand reconstruction is less sensitive to some impaired observation. We also try our method to reconstruct in-the-wild images containing cases including single hand, ego-view, hand-object interaction and truncated hands. Fig. \ref{fig:qua2} presents some representative images where hands are accurately reconstructed, proving that our method has strong generality and is very promising for real-world applications.
\subsection{Ablation study}
\noindent
As introduced in Sec.\ref{Methodology}, our Attention Collaboration-based Feature Aggregator (ACFA) works mainly by collaborating three representations: \textbf{G}lobal representation (G, baseline), \textbf{P}art-based representation (P), and cross-hand-attention prior (C).  Therefore, we investigate the effectiveness of each module. We treat the center-based representation as a baseline and gradually add another module to see their improvement. As shown in Tab. \ref{table:feature}, we can clearly observe both part-based and cross-hand significantly improve the baseline. More interestingly, the improvement of adding C on the IH dataset is more significant than that on the SH dataset. This demonstrates cross-hand-attention prior facilitates addressing interacting hand challenges.
\begin{table}
\centering
\resizebox{0.9\linewidth}{!}{
\centering
\begin{tabular}{c|cccc}
\toprule
 & MPJPE & IH MPJPE  & SH MPJPE   & PAMPJPE \\ 
\midrule \midrule
G(ResNet-50)  & 9.78   & 10.56 & 8.77 & 6.56 \\
G(HRNet-32W)  & 9.56   & 10.35 & 8.65 & 6.41 \\ 
P             & 8.70     & 9.76   & 7.26   & 5.59   \\
G+C           & 9.1  & 9.88 & 8.11   & 6.08  \\
G+P           & 8.52  & 9.69  & 6.87   & 5.49  \\
G+\textbf{C}+P & \textbf{8.09} & \textbf{9.08} & \textbf{6.85} & \textbf{5.21} \\
\hline
\end{tabular}
}
\caption{Ablation study on the part (P), global (G), and cross-hand (C) prior representation. We do not use any extra information such as bounding box and GT scale in ablation study.}
\label{table:feature}
\end{table}

\section{Conclusion and Future Work}\label{Conclusion}
\noindent
%
\textbf{Conclusion:} We present a simple yet effective arbitrary hand reconstruction approach considering more challenges such as interacting hands, truncated hands, and external occlusion from monocular RGB image. To this end, we propose to leverage center and part attention to mitigate interdependencies between hands and between parts to release the input constraint and eliminate the prediction’s sensitivity to a small occluded or truncated part. Experiments show that our method is a promising solution, which can serve as a baseline to inspire more research on arbitrary hand pose and shape reconstruction.

\noindent
\textbf{Limitation \& Future Work:} Our major limitation is the lack of explicit solution for mesh collision, resulting in occasional inter-penetration, which can be solved by leveraging relative information or perspective camera model for accurate depth reasoning and better simulation of translation.

{\small
\bibliographystyle{abbrvnat}
\bibliography{egbib}

\begin{thebibliography}{40}
\providecommand{\natexlab}[1]{#1}
\providecommand{\url}[1]{\texttt{#1}}
\expandafter\ifx\csname urlstyle\endcsname\relax
  \providecommand{\doi}[1]{doi: #1}\else
  \providecommand{\doi}{doi: \begingroup \urlstyle{rm}\Url}\fi

\bibitem[Baek et~al.(2019)Baek, Kim, and Kim]{baek2019pushing}
S.~Baek, K.~I. Kim, and T.-K. Kim.
\newblock Pushing the envelope for rgb-based dense 3d hand pose estimation via
  neural rendering.
\newblock In \emph{Proceedings of the IEEE/CVF Conference on Computer Vision
  and Pattern Recognition}, pages 1067--1076, 2019.

\bibitem[Boukhayma et~al.(2019)Boukhayma, Bem, and Torr]{boukhayma20193d}
A.~Boukhayma, R.~d. Bem, and P.~H. Torr.
\newblock 3d hand shape and pose from images in the wild.
\newblock In \emph{Proceedings of the IEEE/CVF Conference on Computer Vision
  and Pattern Recognition}, pages 10843--10852, 2019.

\bibitem[Cheng et~al.(2020)Cheng, Xiao, Wang, Shi, Huang, and
  Zhang]{cheng2020higherhrnet}
B.~Cheng, B.~Xiao, J.~Wang, H.~Shi, T.~S. Huang, and L.~Zhang.
\newblock Higherhrnet: Scale-aware representation learning for bottom-up human
  pose estimation.
\newblock In \emph{Proceedings of the IEEE/CVF conference on computer vision
  and pattern recognition}, pages 5386--5395, 2020.

\bibitem[Choutas et~al.(2020)Choutas, Pavlakos, Bolkart, Tzionas, and
  Black]{choutas2020monocular}
V.~Choutas, G.~Pavlakos, T.~Bolkart, D.~Tzionas, and M.~J. Black.
\newblock Monocular expressive body regression through body-driven attention.
\newblock In \emph{European Conference on Computer Vision}, pages 20--40.
  Springer, 2020.

\bibitem[Fan et~al.(2021)Fan, Spurr, Kocabas, Tang, Black, and
  Hilliges]{fan2021digit}
Z.~Fan, A.~Spurr, M.~Kocabas, S.~Tang, M.~Black, and O.~Hilliges.
\newblock Learning to disambiguate strongly interacting hands via probabilistic
  per-pixel part segmentation.
\newblock In \emph{International Conference on 3D Vision (3DV)}, 2021.

\bibitem[Feng et~al.(2021)Feng, Choutas, Bolkart, Tzionas, and
  Black]{feng2021collaborative}
Y.~Feng, V.~Choutas, T.~Bolkart, D.~Tzionas, and M.~J. Black.
\newblock Collaborative regression of expressive bodies using moderation.
\newblock In \emph{2021 International Conference on 3D Vision (3DV)}, pages
  792--804. IEEE, 2021.

\bibitem[Ge et~al.(2019)Ge, Ren, Li, Xue, Wang, Cai, and Yuan]{ge20193d}
L.~Ge, Z.~Ren, Y.~Li, Z.~Xue, Y.~Wang, J.~Cai, and J.~Yuan.
\newblock 3d hand shape and pose estimation from a single rgb image.
\newblock In \emph{Proceedings of the IEEE/CVF Conference on Computer Vision
  and Pattern Recognition}, pages 10833--10842, 2019.

\bibitem[He et~al.(2016)He, Zhang, Ren, and Sun]{he2016deep}
K.~He, X.~Zhang, S.~Ren, and J.~Sun.
\newblock Deep residual learning for image recognition.
\newblock In \emph{Proceedings of the IEEE conference on computer vision and
  pattern recognition}, pages 770--778, 2016.

\bibitem[Joo et~al.(2018)Joo, Simon, and Sheikh]{joo2018total}
H.~Joo, T.~Simon, and Y.~Sheikh.
\newblock Total capture: A 3d deformation model for tracking faces, hands, and
  bodies.
\newblock In \emph{Proceedings of the IEEE conference on computer vision and
  pattern recognition}, pages 8320--8329, 2018.

\bibitem[Kato et~al.(2018)Kato, Ushiku, and Harada]{kato2018neural}
H.~Kato, Y.~Ushiku, and T.~Harada.
\newblock Neural 3d mesh renderer.
\newblock In \emph{Proceedings of the IEEE conference on computer vision and
  pattern recognition}, pages 3907--3916, 2018.

\bibitem[Kocabas et~al.(2021)Kocabas, Huang, Hilliges, and
  Black]{kocabas2021pare}
M.~Kocabas, C.-H.~P. Huang, O.~Hilliges, and M.~J. Black.
\newblock Pare: Part attention regressor for 3d human body estimation.
\newblock In \emph{Proceedings of the IEEE/CVF International Conference on
  Computer Vision}, pages 11127--11137, 2021.

\bibitem[Kulon et~al.(2020)Kulon, Guler, Kokkinos, Bronstein, and
  Zafeiriou]{kulon2020weakly}
D.~Kulon, R.~A. Guler, I.~Kokkinos, M.~M. Bronstein, and S.~Zafeiriou.
\newblock Weakly-supervised mesh-convolutional hand reconstruction in the wild.
\newblock In \emph{Proceedings of the IEEE/CVF conference on computer vision
  and pattern recognition}, pages 4990--5000, 2020.

\bibitem[Li et~al.(2022)Li, An, Zhang, Wu, Chen, Yu, and Liu]{Li2022intaghand}
M.~Li, L.~An, H.~Zhang, L.~Wu, F.~Chen, T.~Yu, and Y.~Liu.
\newblock Interacting attention graph for single image two-hand reconstruction.
\newblock In \emph{IEEE/CVF Conf. on Computer Vision and Pattern Recognition
  (CVPR)}, June 2022.

\bibitem[Lin et~al.(2021{\natexlab{a}})Lin, Wang, and Liu]{lin2021end}
K.~Lin, L.~Wang, and Z.~Liu.
\newblock End-to-end human pose and mesh reconstruction with transformers.
\newblock In \emph{Proceedings of the IEEE/CVF Conference on Computer Vision
  and Pattern Recognition}, pages 1954--1963, 2021{\natexlab{a}}.

\bibitem[Lin et~al.(2021{\natexlab{b}})Lin, Wang, and Liu]{lin2021mesh}
K.~Lin, L.~Wang, and Z.~Liu.
\newblock Mesh graphormer.
\newblock In \emph{Proceedings of the IEEE/CVF International Conference on
  Computer Vision}, pages 12939--12948, 2021{\natexlab{b}}.

\bibitem[Lin et~al.(2017)Lin, Goyal, Girshick, He, and
  Doll{\'a}r]{lin2017focal}
T.-Y. Lin, P.~Goyal, R.~Girshick, K.~He, and P.~Doll{\'a}r.
\newblock Focal loss for dense object detection.
\newblock In \emph{Proceedings of the IEEE international conference on computer
  vision}, pages 2980--2988, 2017.

\bibitem[Liu et~al.(2018)Liu, Lehman, Molino, Petroski~Such, Frank, Sergeev,
  and Yosinski]{liu2018intriguing}
R.~Liu, J.~Lehman, P.~Molino, F.~Petroski~Such, E.~Frank, A.~Sergeev, and
  J.~Yosinski.
\newblock An intriguing failing of convolutional neural networks and the
  coordconv solution.
\newblock \emph{Advances in neural information processing systems}, 31, 2018.

\bibitem[Lv et~al.(2021)Lv, Xu, Yang, Qian, Mao, and Lu]{lv2021handtailor}
J.~Lv, W.~Xu, L.~Yang, S.~Qian, C.~Mao, and C.~Lu.
\newblock Handtailor: Towards high-precision monocular 3d hand recovery.
\newblock \emph{arXiv preprint arXiv:2102.09244}, 2021.

\bibitem[Moon and Lee(2020)]{moon2020i2l}
G.~Moon and K.~M. Lee.
\newblock I2l-meshnet: Image-to-lixel prediction network for accurate 3d human
  pose and mesh estimation from a single rgb image.
\newblock In \emph{European Conference on Computer Vision}, pages 752--768.
  Springer, 2020.

\bibitem[Moon et~al.(2020{\natexlab{a}})Moon, Shiratori, and
  Lee]{moon2020deephandmesh}
G.~Moon, T.~Shiratori, and K.~M. Lee.
\newblock Deephandmesh: A weakly-supervised deep encoder-decoder framework for
  high-fidelity hand mesh modeling.
\newblock In \emph{European Conference on Computer Vision}, pages 440--455.
  Springer, 2020{\natexlab{a}}.

\bibitem[Moon et~al.(2020{\natexlab{b}})Moon, Yu, Wen, Shiratori, and
  Lee]{Moon_2020_ECCV_InterHand2.6M}
G.~Moon, S.-I. Yu, H.~Wen, T.~Shiratori, and K.~M. Lee.
\newblock Interhand2.6m: A dataset and baseline for 3d interacting hand pose
  estimation from a single rgb image.
\newblock In \emph{European Conference on Computer Vision (ECCV)},
  2020{\natexlab{b}}.

\bibitem[Mueller et~al.(2019)Mueller, Davis, Bernard, Sotnychenko, Verschoor,
  Otaduy, Casas, and Theobalt]{mueller2019real}
F.~Mueller, M.~Davis, F.~Bernard, O.~Sotnychenko, M.~Verschoor, M.~A. Otaduy,
  D.~Casas, and C.~Theobalt.
\newblock Real-time pose and shape reconstruction of two interacting hands with
  a single depth camera.
\newblock \emph{ACM Transactions on Graphics (TOG)}, 38\penalty0 (4):\penalty0
  1--13, 2019.

\bibitem[Paszke et~al.(2019)Paszke, Gross, Massa, Lerer, Bradbury, Chanan,
  Killeen, Lin, Gimelshein, Antiga, et~al.]{paszke2019pytorch}
A.~Paszke, S.~Gross, F.~Massa, A.~Lerer, J.~Bradbury, G.~Chanan, T.~Killeen,
  Z.~Lin, N.~Gimelshein, L.~Antiga, et~al.
\newblock Pytorch: An imperative style, high-performance deep learning library.
\newblock \emph{Advances in neural information processing systems}, 32, 2019.

\bibitem[Romero et~al.(2017)Romero, Tzionas, and Black]{romero2017embodied}
J.~Romero, D.~Tzionas, and M.~J. Black.
\newblock Embodied hands: Modeling and capturing hands and bodies together.
\newblock \emph{ACM Transactions on Graphics}, 36\penalty0 (6), 2017.

\bibitem[Rong et~al.(2021)Rong, Wang, Liu, and Loy]{rong2021monocular}
Y.~Rong, J.~Wang, Z.~Liu, and C.~C. Loy.
\newblock Monocular 3d reconstruction of interacting hands via collision-aware
  factorized refinements.
\newblock In \emph{2021 International Conference on 3D Vision (3DV)}, pages
  432--441. IEEE, 2021.

\bibitem[Smith et~al.(2020)Smith, Wu, Wen, Peluse, Sheikh, Hodgins, and
  Shiratori]{smith2020constraining}
B.~Smith, C.~Wu, H.~Wen, P.~Peluse, Y.~Sheikh, J.~K. Hodgins, and T.~Shiratori.
\newblock Constraining dense hand surface tracking with elasticity.
\newblock \emph{ACM Transactions on Graphics (TOG)}, 39\penalty0 (6):\penalty0
  1--14, 2020.

\bibitem[Spurr et~al.(2018)Spurr, Song, Park, and Hilliges]{spurr2018cross}
A.~Spurr, J.~Song, S.~Park, and O.~Hilliges.
\newblock Cross-modal deep variational hand pose estimation.
\newblock In \emph{Proceedings of the IEEE Conference on Computer Vision and
  Pattern Recognition}, pages 89--98, 2018.

\bibitem[Sun et~al.(2021)Sun, Bao, Liu, Fu, Michael~J., and Mei]{ROMP}
Y.~Sun, Q.~Bao, W.~Liu, Y.~Fu, B.~Michael~J., and T.~Mei.
\newblock Monocular, one-stage, regression of multiple 3d people.
\newblock In \emph{ICCV}, 2021.

\bibitem[Tang et~al.(2021)Tang, Wang, and Fu]{tang2021towards}
X.~Tang, T.~Wang, and C.-W. Fu.
\newblock Towards accurate alignment in real-time 3d hand-mesh reconstruction.
\newblock In \emph{Proceedings of the IEEE/CVF International Conference on
  Computer Vision}, pages 11698--11707, 2021.

\bibitem[Taylor et~al.(2017)Taylor, Tankovich, Tang, Keskin, Kim, Davidson,
  Kowdle, and Izadi]{taylor2017articulated}
J.~Taylor, V.~Tankovich, D.~Tang, C.~Keskin, D.~Kim, P.~Davidson, A.~Kowdle,
  and S.~Izadi.
\newblock Articulated distance fields for ultra-fast tracking of hands
  interacting.
\newblock \emph{ACM Transactions on Graphics (TOG)}, 36\penalty0 (6):\penalty0
  1--12, 2017.

\bibitem[Wang et~al.(2020)Wang, Mueller, Bernard, Sorli, Sotnychenko, Qian,
  Otaduy, Casas, and Theobalt]{wang2020rgb2hands}
J.~Wang, F.~Mueller, F.~Bernard, S.~Sorli, O.~Sotnychenko, N.~Qian, M.~A.
  Otaduy, D.~Casas, and C.~Theobalt.
\newblock Rgb2hands: real-time tracking of 3d hand interactions from monocular
  rgb video.
\newblock \emph{ACM Transactions on Graphics (ToG)}, 39\penalty0 (6):\penalty0
  1--16, 2020.

\bibitem[Xiang et~al.(2019)Xiang, Joo, and Sheikh]{xiang2019monocular}
D.~Xiang, H.~Joo, and Y.~Sheikh.
\newblock Monocular total capture: Posing face, body, and hands in the wild.
\newblock In \emph{Proceedings of the IEEE/CVF conference on computer vision
  and pattern recognition}, pages 10965--10974, 2019.

\bibitem[Zhang et~al.(2021{\natexlab{a}})Zhang, Wang, Deng, Zhang, Tan, Ma, and
  Wang]{zhang2021interacting}
B.~Zhang, Y.~Wang, X.~Deng, Y.~Zhang, P.~Tan, C.~Ma, and H.~Wang.
\newblock Interacting two-hand 3d pose and shape reconstruction from single
  color image.
\newblock In \emph{Proceedings of the IEEE/CVF International Conference on
  Computer Vision}, pages 11354--11363, 2021{\natexlab{a}}.

\bibitem[Zhang et~al.(2019)Zhang, Li, Mo, Zhang, and Zheng]{zhang2019end}
X.~Zhang, Q.~Li, H.~Mo, W.~Zhang, and W.~Zheng.
\newblock End-to-end hand mesh recovery from a monocular rgb image.
\newblock In \emph{Proceedings of the IEEE/CVF International Conference on
  Computer Vision}, pages 2354--2364, 2019.

\bibitem[Zhang et~al.(2021{\natexlab{b}})Zhang, Li, An, Li, Yu, and
  Liu]{zhang2021lightweight}
Y.~Zhang, Z.~Li, L.~An, M.~Li, T.~Yu, and Y.~Liu.
\newblock Lightweight multi-person total motion capture using sparse multi-view
  cameras.
\newblock In \emph{Proceedings of the IEEE/CVF International Conference on
  Computer Vision}, pages 5560--5569, 2021{\natexlab{b}}.

\bibitem[Zhou et~al.(2019)Zhou, Barnes, Lu, Yang, and Li]{zhou2019continuity}
Y.~Zhou, C.~Barnes, J.~Lu, J.~Yang, and H.~Li.
\newblock On the continuity of rotation representations in neural networks.
\newblock In \emph{Proceedings of the IEEE/CVF Conference on Computer Vision
  and Pattern Recognition}, pages 5745--5753, 2019.

\bibitem[Zhou et~al.(2020)Zhou, Habermann, Xu, Habibie, Theobalt, and
  Xu]{zhou2020monocular}
Y.~Zhou, M.~Habermann, W.~Xu, I.~Habibie, C.~Theobalt, and F.~Xu.
\newblock Monocular real-time hand shape and motion capture using multi-modal
  data.
\newblock In \emph{Proceedings of the IEEE/CVF Conference on Computer Vision
  and Pattern Recognition}, pages 5346--5355, 2020.

\bibitem[Zhou et~al.(2021)Zhou, Habermann, Habibie, Tewari, Theobalt, and
  Xu]{zhou2021monocular}
Y.~Zhou, M.~Habermann, I.~Habibie, A.~Tewari, C.~Theobalt, and F.~Xu.
\newblock Monocular real-time full body capture with inter-part correlations.
\newblock In \emph{Proceedings of the IEEE/CVF Conference on Computer Vision
  and Pattern Recognition}, pages 4811--4822, 2021.

\bibitem[Zimmermann and Brox(2017)]{zb2017hand}
C.~Zimmermann and T.~Brox.
\newblock Learning to estimate 3d hand pose from single rgb images.
\newblock Technical report, arXiv:1705.01389, 2017.
\newblock URL \url{https://lmb.informatik.uni-freiburg.de/projects/hand3d/}.
\newblock https://arxiv.org/abs/1705.01389.

\bibitem[Zimmermann et~al.(2019)Zimmermann, Ceylan, Yang, Russell, Argus, and
  Brox]{zimmermann2019freihand}
C.~Zimmermann, D.~Ceylan, J.~Yang, B.~Russell, M.~Argus, and T.~Brox.
\newblock Freihand: A dataset for markerless capture of hand pose and shape
  from single rgb images.
\newblock In \emph{Proceedings of the IEEE/CVF International Conference on
  Computer Vision}, pages 813--822, 2019.

\end{thebibliography}
}

\end{document}